# SSMT: Few-Shot Traffic Forecasting with Single Source Meta-Transfer

Kishor Kumar Bhaumik[1,3] Minha Kim[1] Fahim Faisal Niloy[2] Amin Ahsan Ali[3] Simon S. Woo[1]★

[1] Sungkyunkwan University, South Korea
{kishor25,sunshine01,swoo}@g.skku.ediu
[2] University of California, Riverside
fnilo001@ucr.edu
[3] Center for Computational & Data Sciences, Independent University, Bangladesh
aminali@iub.edu.bd

**Abstract.** Traffic forecasting in Intelligent Transportation Systems (ITS) is vital for intelligent traffic prediction. Yet, ITS often relies on data from traffic sensors or vehicle devices, where certain cities might not have all those smart devices or enabling infrastructures. Also, recent studies have employed meta-learning to generalize spatial-temporal traffic networks, utilizing data from multiple cities for effective traffic forecasting for data-scarce target cities. However, collecting data from multiple cities can be costly and time-consuming. To tackle this challenge, we introduce **S**ingle **S**ource **M**eta-**T**ransfer Learning (*SSMT*) which relies only on a single source city for traffic prediction. Our method harnesses this transferred knowledge to enable few-shot traffic forecasting, particularly when the target city possesses limited data. Specifically, we use memory-augmented attention to store the heterogeneous spatial knowledge from the source city and selectively recall them for the data-scarce target city. We extend the idea of sinusoidal positional encoding to establish meta-learning tasks by leveraging diverse temporal traffic patterns from the source city. Moreover, to capture a more generalized representation of the positions we introduced a meta-positional encoding that learns the most optimal representation of the temporal pattern across all the tasks. We experiment on five real-world benchmark datasets to demonstrate that our method outperforms several existing methods in time series traffic prediction. Our code is available at https://github.com/Kishor-Bhaumik/SSMT.

## 1 Introduction

Accurate traffic forecasting is crucial for Intelligent Transportation Systems (ITS) to enable a wide range of AI services that rely on real time traffic information, such as food delivery, taxi services, etc. Traditional methods such as ARIMA [1] and Kalman filter often rely on historical data for univariate time series forecasting. Lately, spatiotemporal traffic forecasting methods [32,27,9] are proposed to integrate temporal and topological sensor relationships for multivariate traffic forecasting using Graph Neural Networks (GNNs) [27]. However, such methods generally rely on abundant training data and fail to generalize for the data-scarce cities.

---

★ Corresponding Author

Recently, transfer learning-based methods such as RegionTrans [24] and Cross TRes [12] focused on grid-based traffic prediction by transferring source knowledge from multiple source cities that have abundant data to the target cities that have limited data. They include large-scale auxiliary data to better match regions that are similar to one another. However, these methods are not directly applicable to graph based traffic prediction because their grid structure differs significantly from the graph structure of the traffic network in target cities. To tackle this problem, few transferable time series forecasting models have been proposed. MetaST [26] uses a global learnable memory, while ST-MetaNet [18] and ST-GFSL [17] employ meta-knowledge from multiple cities for graph-based traffic forecasting. However, the aforementioned approaches tend to overlook the challenges and disadvantages of collecting data from diverse cities. Indeed, gathering sensor data from multiple cities can be prohibitively expensive. Also, possibly, negative transfer [30], when an underdeveloped city or the city significantly differs from the target city is included in the source data, can further undesirably decrease the forecasting performance for the target city. Addressing the aforementioned prior challenges, we introduce Few-Shot Traffic Forecasting with Single Source Meta-Transfer Learning ($SSMT$), a novel framework for transferable spatiotemporal traffic prediction. Our method utilizes data from a single source city learning important spatial footprints and effectively adapting to new temporal patterns. Overall, the main contributions of our work are summarized as follows:

– We present $SSMT$, a new meta-learning based time series traffic forecasting approach that allows single source knowledge transfer from source city to the target city. We carefully construct three meta-learning tasks to help the model quickly adapt to different temporal resolution, each based on a particular periodic pattern (daily, weekly, or monthly).

– Our $SSMT$ employs an external memory module to facilitate the effective transfer of topological information from source to target cities in the presence of sensor node discrepancy. This allows the model to gain a deeper understanding of the spatial context, resulting in more accurate predictions for the target city.

– We empirically demonstrate the effectiveness of our proposed $SSMT$ framework on five real-world spatiotemporal datasets and show that our method achieves superior performances compared to the existing baselines.

## 2   Related Work

With the emergence of deep learning and graph neural networks, graph is applied to tackle a variety of urban challenges to explore spatial structural interactions. Andrea et al. [3] proposed a framework to explain empirical results associated with the use of trainable node embeddings and discuss different architectures and regularization techniques to account for local effects. Bai et al. [2] present STG2Seq, a graph-based model for multi-step citywide passenger demand. Yuan et al. [29] proposed recasting spatio-temporal few-shot learning as pre-training a generative diffusion model, which creates tailored neural networks guided by prompts. This approach allows for adaptability to diverse data distributions and city-specific characteristics. To capture the dynamic aspects of urban traffic flow, Lu et al. [16] propose spatial and semantic neighbors of road segments. Do et al. [5] use IoT sensors on automobiles to assess city air quality and use variational graph autoencoders to predict unknown air pollutants. Nevertheless, these approaches are non-transferable and primarily concentrate on single-city traffic forecasting. Few-shot learning (FSL) has shown promising performance in various domains such as computer vision, natural language processing, and reinforcement learning when dealing with data scarcity. In our context, when data-rich source cities are used to transfer knowledge to data-scarce target cities, this problem is referred to as few-shot traffic forecasting. And, recently cross-city transfer learning models [24,12] have gained significant popularity in this area. MetaST [26] employs a global memory queried by the target region. Moreover, STrans-GAN [31] generates future traffic speed using GANs, and TPB method [15] proposes a traffic pattern bank to store similar patterns from multiple source cities for the downstream fine-tuning task. However, these methods heavily depend on data-rich multiple source cities, and can be cost-prohibitive in practice.

Recently, memory-augmented attention (MAA) models have gained much attention for capturing long-term dependency, particularly anomaly detection task [8]. Park et al.[19] used MAA for anomaly detection in video sequences. These studies explicitly utilize memory augmented attention to enhance model performance by storing data patterns. Inspired by these, in our work, we propose a separate memory module to address non-transferability due to node count mismatches between source and target cities.

## 3   Methodology

### 3.1   Preliminaries and Problem Formulation

Traffic networks are represented as $\mathcal{G}_s$ and $\mathcal{G}_t$, with $\mathcal{V}$ as vertices (e.g., traffic sensors) and $\mathcal{E}$ as edges (connectivity). $V$ is the set of vertices where $V \in (v_1, v_2, v_3, ...v_n)$. In our context, we use subscripts $s$ and $t$ to denote the data of the source and target city, respectively. Next, we define the source data as $X_s \in \mathbb{R}^{N_s \times T_s \times C_s}$ and target data as $X_t \in \mathbb{R}^{N_t \times T_t \times C_t}$, where $N$, $T$, and $C$ represents the total number of nodes, time window length, and the number of traffic features, respectively. The adjacency matrix $\mathcal{A} \in \mathbb{R}^{N \times N}$ is the spatiotemporal graph of $\mathcal{G}$. $A_{ij} = 1$ indicates that there is an edge between node $v_i$ and $v_j$, otherwise, $v_{ij} = 0$. In our work, we mainly focus on investigating the transferability of a single feature, namely traffic speed, thereby reducing $C$ to 1. Hence, we finally consider $X_s \in \mathbb{R}^{N_s \times T_s}$ as the source, and $X_t \in \mathbb{R}^{N_t \times T_t}$ as the target data. Also, as the number of nodes differs between the source and the target city, we denote $N_s \neq N_t$ to indicate this discrepancy in the number of nodes.

In general, the transferable traffic forecasting problem can be divided into two stages: 1) pre-training and 2) fine-tuning. First, in the pre-training stage, we can formulate the forecasting problem by training a mapping function $f_{\theta_s}$ on the source data, which predicts future timestamps ($T'$) based on past timestamps ($T$). We denote the historical spatiotemporal input data and the predicted observations as $\langle X_{\mathcal{G}_s}^{(t-T+1)}, X_{\mathcal{G}_s}^{(t-T+2)}, \ldots, X_{\mathcal{G}_s}^{(t)} \rangle$ and $\langle X_{\mathcal{G}_s}^{(t+1)}, X_{\mathcal{G}_s}^{(t+2)}, \ldots, X_{\mathcal{G}_s}^{(t+T')} \rangle$, respectively. Thus, the time series forecasting with the input graph $\mathcal{G}_s$ for the source city can be defined as follows:

$$\langle X_{\mathcal{G}_s}^{(t-T+1)}, X_{\mathcal{G}_s}^{(t-T+2)}, \ldots, X_{\mathcal{G}_s}^{(t)} \rangle \xrightarrow{f_{\theta_s}} \langle X_{\mathcal{G}_s}^{(t+1)}, X_{\mathcal{G}_s}^{(t+2)}, \ldots, X_{\mathcal{G}_s}^{(t+T')} \rangle$$

Next, in the fine-tuning stage, the forecasting task is performed by fine-tuning the same mapping function with parameters $\theta_s$, which are initially shared from the pre-trained function. The fine-tuning process aims to improve the model's ability to predict graph signals specifically for the target road network. And, the fine-tuning process can be defined as follows:

$$\langle X_{\mathcal{G}_t}^{(t-T+1)}, X_{\mathcal{G}_t}^{(t-T+2)}, \ldots, X_{\mathcal{G}_t}^{(t)} \rangle \xrightarrow{f_{\theta_t^*;\theta_s}} \langle X_{\mathcal{G}_t}^{(t+1)}, X_{\mathcal{G}_t}^{(t+2)}, \ldots, X_{\mathcal{G}_t}^{(t+T')} \rangle$$

In particular, the notation $f_{\theta_t^*;\theta_s}$ represents the adjusted function parameters from $\theta_s$ to fit the target domain. It basically signifies the updated or adapted set of parameters that have been modified from the source model to better align with the characteristics and requirements of the target city.

### 3.2 Meta-Learning Framework

In our model, $\theta$ represents the encoded spatiotemporal model parameter for the traffic prediction. To adapt the source model parameters effectively to a target city, we adopt the approach proposed in model-agnostic meta-learning (MAML) [7] In MAML framework, the dataset is divided into multiple tasks and each task is divided further into support and query sets. Then, the support set is used for model adaptation during meta-training, while the query set evaluates the model's performance after this adaptation. We use MAML to initialize $\theta$ with multiple tasks from the source city as $\theta_s$, such that it can minimize the average generalization loss across all source tasks. When adapting to a new task $\mathcal{T}^i$, the model's parameters $\theta_s$ are updated to $\theta_i'$. In MAML, the inner loop is for training the model's parameters for a specific task, while the outer loop is to adjust the model's initial parameters to enhance the learning performance

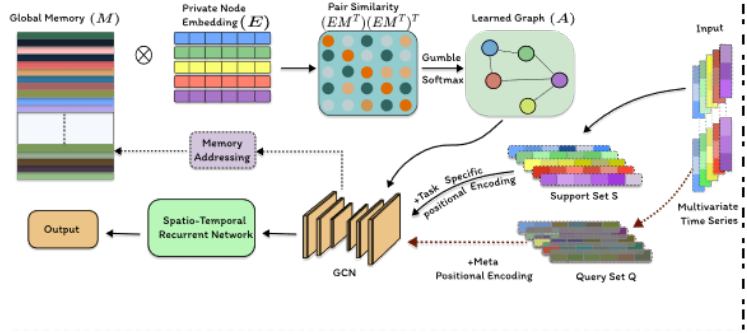

Fig. 1: The overall architecture of our proposed *SSMT* consists of several components. Firstly, the discrete graph $A$ is learned by computing the pair similarity between $M$ and $E$. The memory component then addresses the GCN module's output, which is subsequently delivered to the spatiotemporal recurrent block. Finally, the output of the spatiotemporal block is utilized to compute the loss.

across a range of tasks. The inner loop optimization of MAML can be expressed as follows:

$$\theta'_i = \theta_s - \alpha \nabla_\theta \mathcal{L}_{\mathcal{T}^i}(f_{\theta_s}) \tag{1}$$

And, practically, we can perform multiple steps of gradient descent to update the initialization $\theta_s$ to $\theta'_i$. For each task $\mathcal{T}^i$, the training process is iterated on batches of tasks sampled from source tasks $S(\mathcal{T})$. More intuitively, the meta-learning objective is defined as follows:

$$\min_{\theta_s} \sum_{\mathcal{T}^i \sim S(\mathcal{T})} \mathcal{L}_{\mathcal{T}^i}(f_{\theta'_i}) = \sum_{\mathcal{T}^i \sim S(\mathcal{T})} \mathcal{L}_{\mathcal{T}^i}\left(f_{\theta_s - \alpha \nabla_\theta \mathcal{L}_{\mathcal{T}^i}(f_{\theta_s})}\right) \tag{2}$$

Finally, the outer loop optimization across all the tasks can be defined as:

$$\theta = \theta - \beta \nabla_{\theta'_i} \sum_{\mathcal{T}^i \sim S(\mathcal{T})} \mathcal{L}_{\mathcal{T}^i}(f_{\theta'_i}) \tag{3}$$

Defining $\beta$ as the meta-step size, we assume that $\theta$ will yield superior generalization performance to a target city, as it provides an initialization that performs well across multiple tasks.

### 3.3 Positional-Encoding Driven Task Partitioning

Let us define each batch of input data as $D_j$ where $j \in [1, 2, 3, .., BatchSize]$. We first represent this split of a single batch in three distinct sets to present daily, weekly and monthly periodic patterns as follows:

$$D_j = \mathcal{B}_j^{(1)} \cup \mathcal{B}_j^{(2)} \cup \mathcal{B}_j^{(3)} \tag{4}$$
$$\mathcal{B}_j^i = \{(x_1^{(i)}, y_1^{(i)}), (x_2^{(i)}, y_2^{(i)}), \ldots, (x_{n^{(i)}}^{(i)}, y_{n^{(i)}}^{(i)})\}$$

where $\mathcal{B}^i$ represents the $i$-th set of the batch and $i \in [1, 2, 3]$. $\mathcal{B}^i$ contains multiple data points and each data point $(x_j^{(i)}, y_j^{(i)})$ consists of input features $x_j^{(i)}$ and corresponding labels $y_j^{(i)}$ for the $i$-th set. $n^{(i)}$ is the number of data points in the $i$-th set of the batch. Inspired by the widely adopted concept of relative positional encoding in transformer-based attention mechanisms [22,13], we derive positional encoding as follows:

$$PE_{\text{pos},2k} = \sin\left(\frac{2\pi \cdot \text{pos}}{24 \cdot \text{samples\_per\_hour} \cdot V}\right)$$
$$PE_{\text{pos},2k+1} = \cos\left(\frac{2\pi \cdot \text{pos}}{24 \cdot \text{samples\_per\_hour} \cdot V}\right), \tag{5}$$

where the value of $V$ is influenced by the periodic patterns observed in the source city with $k$ representing specific positions (either odd or even). For instance, to capture the daily pattern, we set $V = 1$. Similarly, for the weekly pattern, $V$ is set to 7, and for the monthly pattern, $V$ is set to 30. We next add these positional encodings to each of $\mathcal{B}^i$ resulting in three unique sets of batches becoming three distinct tasks. We then equally divide each task into support and query sets for the inner loop optimization.

However, traditional sinusoidal positional encodings might not be optimal to use for all tasks. To address this challenge, we further introduce meta-positional encoding that is learned throughout the outer loop optimization process, denoted as $\eta \in \mathbb{R}^{N \times T}$. We define the meta-positional encoding as follows:

$$\eta = \alpha \otimes E; \text{ where, } \alpha \in \mathbb{R}^{1 \times T} \text{ and } E \in \mathbb{R}^{N \times T}, \tag{6}$$

where $\otimes$ represents the matrix multiplication, $\alpha$ is the scaling parameter and $E$ is the embedding vector. And, the scaling parameter, $\alpha$, acts as a weight that determines the importance of the positional information. The meta-positinal encoding $\eta$ remains static in the inner loop and is only updated during the outer loop optimization. By learning a positional encoding in the outer loop, our model can potentially capture a more generalized representation of position that works across different time scales (daily, weekly, and monthly). Furthermore, it can help the meta-model start with a more suitable positional representation for a variety of tasks before inner loop adaptation.

### 3.4  Memory-based Spatial Knowledge Transfer

While traditional methods [28,11] generate graphs from a similarity matrix, our transfer-learning context for inter-city traffic forecasting presents the following unique challenges: When transferring knowledge from a source city to a target city, node embeddings become a problem due to discrepancies in the number of sensors between the two cities. For example, let us consider a scenario where the model is pre-trained using the source data, resulting in the learned node embedding $E_s \in \mathbb{R}^{N_s \times d}$. However, creating a target city specific graph requires the node embedding $E_t \in \mathbb{R}^{N_t \times d}$ which disables to use of the learnable node embedding from the source city directly.

In order to overcome this non-transferability issue due to the mismatch of node number between source and target cities, we adopt a memory module inspired by [23,21]. And, our proposed module supplements node embeddings that cannot be directly transferred. Let $E_s \in \mathbb{R}^{N_s \times d}$ and $E_t \in \mathbb{R}^{N_t \times d}$ denote the private node embeddings of the source and the target, respectively, which are not shared but trained independently. In particular, we leverage the idea of memory network [8] and learn a global memory that is shared in both the source and target city. Specifically, the memory is defined as $M \in \mathbb{R}^{b \times d}$, where $b$ and $d$ denote the number of memory items and the dimension of each item, respectively. During the pre-training step for the source data, the inter-node similarity matrix $\xi_s \in \mathbb{R}^{N_s \times N_s}$ is defined as follows:

$$\xi_s = (E_s M^T)(E_s M^T)^T \qquad (7)$$

Upon completion of the pre-training using the source data, we use only the learned global memory $M$ in the fine-tuning stage for the target data. Likewise, the inter-node similarity matrix $\xi_t \in \mathbb{R}^{N_t \times N_t}$ for the target city can be formulated as follows:

$$\xi_t = (E_t M^T)(E_t M^T)^T, \qquad (8)$$

where the global memory $M$ enables us to transfer the extracted knowledge on the road topological structure of the source data, which could have been potentially lost due to the disparity in the number of nodes between the cities.

To effectively ensure a decent level of sparsity of the graph structure, we apply the Gumbel softmax trick to retrieve the final sparse adjacency matrix $A \in \mathbb{R}^{N \times N}$ for both source and target, where $\sigma$ and $\tau$ are the activation function and the temperature variable, respectively. This process can be expressed as follows:

$$\begin{aligned} A &= \sigma((log(\xi_{ij}/(1-\xi_{ij}) + (n_{ij}^1 - n_{ij}^2))/\tau) \\ s.t.\ & n_{ij}^1, n_{ij}^2 \sim Gumbel(0,1) \end{aligned} \qquad (9)$$

Equation 9 implements the Gumbel Softmax algorithm [10] for our task, where $A_{i,j} = 1$ with the probability $\xi_{i,j}$ and 0 with the remaining probability. Gumbel Softmax maintains the same probability distribution as the normal Softmax, ensuring statistical consistency in generating the trainable probability matrix for the graph forecasting network. And, let $I_n$ denote an identity matrix and $D$ represent a diagonal degree matrix satisfying $D_{ii} = \Sigma_j A_{ij}$, then the specific operation of graph convolutional network ($GCN$) can be expressed as

follows:

$$O = GCN_{\star A}(X) = W(I_n + D^{-\frac{1}{2}} A D^{-\frac{1}{2}})X, \tag{10}$$

where $GCN$ is parameterized by $W \in \mathbb{R}^{T \times d}$, and the output of GCN is denoted as $O \in \mathbb{R}^{N \times d}$.

### 3.5 Memory Addressing

The memory $M$ is designed to explicitly record the topological pattern during the training. We define the memory that computes attention weights $w$ based on the similarity of the memory items and the query from the GCN output $O$. We compute each $w_j$ via a softmax operation as a memory addressing scheme by following:

$$w_j = \frac{exp(sim(O_t^{(a)}, M_j))}{\Sigma_{j=1}^{M} exp(sim(O_t^{(a)}, M_j))} \tag{11}$$

$$sim\left(O_t^{(a)}, M_j\right) = \frac{O_t^{(a)} M_j^\top}{\|O_t^{(a)}\| \, \|M_j\|} \tag{12}$$

where we denote $a$ as a row index, and derive the memory reading operation by matching $O^{(a)}$ with each memory $M_j$. We then calculate a scaler $w_j$ that represents the cosine similarity between vector $O^{(a)}$ and memory $M_j$ as shown in Eq. 12. After that, We recover a node embedding vector $P_t^{(a)} \in \mathbb{R}^d$ combining all the memory item as follows:

$$P_t^{(a)} = \Sigma_{j=1}^{M} w_j m_j \tag{13}$$

### 3.6 Spatial Temporal Recurrent Network

To capture the spatiotemporal pattern in the traffic data, we use the Spatiotemporal Recurrent Graph Convolution Module (STRGC) proposed in [14]. This module integrates node embedding $P_t$ with a Gated Recurrent Unit (GRU) network and processes the input series $X_{t-T+1:t}$ and matrix $P_t$ to ultimately produce the predicted future traffic data, $X_{t+1:t+T'}$. This prediction process is precisely defined as follows:

$$\begin{aligned}
z_t &= \sigma(P_t([X_{0:t} \,||\, h_{t-1}])) \\
r_t &= \sigma(P_t([X_{0:t} \,||\, h_{t-1}])) \\
c_t &= tanh(P_t([X_{0:t} \,||\, (r_t \odot h_{t-1})])) \\
h_t &= z_t \odot h_{t-1} + (1 - z_t) \odot c_t.
\end{aligned} \tag{14}$$

In particular, Figure 1 presents the overall architecture of our method, where we use mean absolute error (MAE) as our objective criterion which is formulated

as follows:

$$\mathcal{L}_{MAE} = \frac{1}{T_y} \sum_{i=1}^{T_y} \|Y_{[:,\tau]} - \widehat{Y}_{[:,\tau]}\|, \qquad (15)$$

To further enhance our model's discriminating power for diverse scenarios on different roads over time, we regulate the memory parameters with two constraints [20] including a contrastive loss $\mathcal{L}_{\text{separate}}$ and a consistency loss $\mathcal{L}_{\text{compact}}$, as follows:

$$\mathcal{L}_{\text{separate}} = \sum_{t}^{T} \sum_{a}^{N} \left[ \left\| O_t^{(a)} - M_p \right\|_2 - \left\| O_t^{(a)} - M_n \right\|_2 + \lambda \right]_+$$

$$\mathcal{L}_{\text{compact}} = \sum_{t}^{T} \sum_{a}^{N} \left\| O_t^{(a)} - M_p \right\|_2$$

where $T$ indicates the total number of sequences (i.e., samples) in the training set, and $p, n$ signify the top two memory item indices determined by ranking $w_j$ in Eq. 11 given localized query $O_t^{(a)}$. And, we regard $O_t^{(a)}$ as the anchor, its most comparable prototype $M_p$ as the positive sample, and the second similar prototype $M_n$ as the negative sample by applying these two constraints, where $\lambda$ signifies the margin between the positive and negative pairings. Here, the idea is to keep memory items as compact as possible, at the same time, as dissimilar as possible through contrastive loss. These two competing objectives limits the memory's ability to directly discriminate between diverse spatiotemporal patterns at the node level. In practice, we found that including them within the objective criteria (i.e., MAE) promotes training convergence (with balancing factors $C_1$, $C_2$ and $C_3$):

$$\mathcal{L}_{total} = C_1 \mathcal{L}_{MAE} + C_2 \mathcal{L}_{\text{separate}} + C_3 \mathcal{L}_{\text{compact}} \qquad (16)$$

It should be noted that during source pre-training, we use only the MAE loss Eq. 15 to update the model parameters. And, for fine-tuning in the target dataset, we employ Eq. 16 as the loss function. This is because we expect the memory to include a variety of topological patterns from the source city. Since data is limited in the target city, the memory should only retain certain patterns related to its own topological structure in order to prevent negative transfer.

### 3.7 Pre-training and Fine-tuning Process

As mentioned, the training process for *SSMT* consists of two primary stages: 1) pre-training with the data from the source city and 2) subsequent fine-tuning using the data from the target city. In the pre-training stage, data from each task is divided into support and query sets to facilitate MAML training. During the inner loop optimization, the model parameters are updated using the support set for each task. Importantly, the learnable meta-positional encoding used for the query set remains static during the inner loop optimization. In the subsequent outer loop optimization phase, the meta-positional encoding is updated based on parameters learned from all the tasks. This enables our model to learn task-universal representations across all temporal patterns. When the model undergoes the fine-tuning stage with the target data, we use the pre-trained weights as the initial parameters, except for the node embedding for target data which is initialized separately. The entire learning process is presented in Algorithm 1

## 4 Experiments

### 4.1 Datasets

We perform our experiments on five publicly available benchmark traffic datasets: METR-LA, PEMS-BAY, PEMSD4, Didi-Chengdu and Didi-Shenzhen [14,4]. These datasets contain months of traffic speed data. METR-LA and PEMS-BAY are collected every five minutes, while Didi-Chengdu and Didi-Shenzhen are collected every ten minutes. In our experiment, we use the PEMS-BAY dataset as the source dataset for the target METR-LA and PEMSD4 datasets, and we use the Didi-Shenzhen dataset as the source dataset for the target Didi-Chengdu dataset since PEMS-BAY and Didi-Shenzhen have significantly more traffic data compared to their respective target cities. Our primary motivation in this paper is to explore single-source transfer learning. Therefore, we strategically selected the largest datasets from source cities to ensure robustness. We assume that target cities will have smaller datasets, as they may have recently deployed sensors. Additionally, we restricted our experiments to cities within the same country due to the high security and privacy concerns associated with traffic data.

Table 1: Dataset description.

| Dataset | Number of sensors (Nodes) | Time Steps | Mean |
|---|---|---|---|
| PEMS-BAY | 325 | 52,116 | 61.77 |
| METRA-LA | 207 | 34,272 | 58.27 |
| PEMSD4 | 170 | 17,856 | 35.38 |
| Didi-Shenzhen | 627 | 17,280 | 31.01 |
| Didi-Chengdu | 524 | 17,280 | 29.02 |

### 4.2 Few-Shot Setting

We use the similar few-shot traffic forecasting setting proposed in [17]. We divide the data of the cities into source, target, and test sets, where the source data consists of data from a single city, while the target and test data consist of data from the target city. For example, if PEMSD4 is the target city, we use the full PEMS-BAY dataset as the source data, 1 week of PEMSD4 data as the target data, and the remaining PEMSD4 data as the test data. We pre-train and learn multiple tasks on the source data and fine-tune the model on the target

SSMT: Few-Shot Traffic Forecasting with Single Source Meta-Transfer        11

---

**Algorithm 1** :Pre-training and fine-tuning process of $SSMT$

---

**Input:** Source input data$(\mathcal{X}^s, \mathcal{Y}^s)$ and target input data $(\mathcal{X}^t, \mathcal{Y}^t)$
**Output:** Spatial temporal prediction in the target city

/* ─────── Source pre-training ─────── */
1: **randomly initialize** $\theta$
2: $\theta_s \longleftarrow \theta$
3: **while** not done **do**
4:     // sample batch from the source dataset
5:     $D_j^s \longleftarrow SampleBatch(\mathcal{X}^s, \mathcal{Y}^s)$
6:     **for all** $D_j^s$ **do**
7:         // sample task from a single batch
8:         $\mathcal{T}^i \longleftarrow D_j^s$ by Eq. (4)
9:         **for all** $\mathcal{T}^i$ **do**
10:            // sample support set
11:            $(\mathcal{X}_{sup}^s, \mathcal{Y}_{sup}^s) \longleftarrow \mathcal{T}^i$
12:            calculate $\alpha \nabla_\theta \mathcal{L}_{\mathcal{T}^i}\left(\mathcal{Y}_{supp}^s, \hat{\mathcal{Y}}_{supp}^s\right)$
                 by Eq. 15
13:            $\theta_s \longleftarrow \theta_s - \alpha \nabla_\theta \mathcal{L}_{\mathcal{T}^i}\left(\mathcal{Y}_{supp}^s, \hat{\mathcal{Y}}_{supp}^s\right)$
14:        **end for**
15:        sample query set $(\mathcal{X}_{qu}^s, \mathcal{Y}_{qu}^s) \longleftarrow \mathcal{T}^i$
16:        compute $\mathcal{L}_{\mathcal{T}^i}\left(\mathcal{Y}_{qu}^s, \hat{\mathcal{Y}}_{qu}^s\right)$ by Eq. 15
17:        $\theta \longleftarrow \beta \nabla_{\theta_s} \sum_{\mathcal{T}^i \sim D_j^s} \mathcal{L}_{\mathcal{T}_i}\left(\mathcal{Y}_{qu}^s, \hat{\mathcal{Y}}_{qu}^s\right)$
18:    **end for**
19: **end while**
/* ─────── Target fine-tuning ─────── */
20: **while** not done **do**
21:    // sample batch from the target dataset
22:    $D_j^t \longleftarrow SampleBatch(\mathcal{X}^t, \mathcal{Y}^t)$
23:    **for all** batches **do**
24:        $(\mathcal{X}_i^t, \mathcal{Y}_i^t) \longleftarrow D_j^t$
25:        $\hat{\mathcal{Y}}_i^t \longleftarrow f_\theta(\mathcal{X}_i^t)$
26:        calculate $\mathcal{L}_{total}\left(\mathcal{Y}_i^t, \hat{\mathcal{Y}}_i^t\right)$ by Eq (16)
27:        calculate $\nabla_\theta \mathcal{L}_{total}\left(\mathcal{Y}_i^t, \hat{\mathcal{Y}}_i^t\right)$ by Eq (16)
28:        $\theta \longleftarrow AdamOptimizer\left(\nabla_\theta \mathcal{L}_{total}\left(\mathcal{Y}_i^t, \hat{\mathcal{Y}}_i^t\right)\right)$
29:    **end for**
30: **end while**
31: **return** $\theta$

---

data. Finally, we evaluate our framework on the test data. We implemented our model using Pytorch training with a single NVIDIA RTX A5000 GPU with 24 GB memory. The training setup involved a batch size of 64 and a maximum of 100 epochs. We set the learning rates for the inner and outer loops at 0.01 and 0.001, respectively. And the memory module is comprised of 20 units, each with a dimension of 64. We empirically chose the values of $C_1$, $C_2$, and $C_3$ as 0.5, 0.2, and 0.3, respectively for Eq. 16.

Table 2: The overall performance of *SSMT* vs. baseline methods, where our method achieves the best performance.

| Methods | PEMS-BAY to METRA-LA | | | | | | PEMS-BAY to PEMSD4 | | | | | |
|---|---|---|---|---|---|---|---|---|---|---|---|---|
| | MAE | | | RMSE | | | MAE | | | RMSE | | |
| | 5 min | 15 min | 30 min | 5 min | 15 min | 30 min | 5 min | 15 min | 30 min | 5 min | 15 min | 30 min |
| DCRNN | 3.05 | 3.45 | 4.38 | 4.78 | 5.97 | 7.55 | 19.55 | 20.56 | 21.15 | 29.19 | 30.24 | 32.33 |
| GWN | 3.12 | 3.59 | 4.27 | 4.87 | 5.94 | 7.66 | 19.37 | 20.58 | 21.12 | 29.26 | 30.15 | 32.24 |
| AdaRNN | 2.95 | 3.39 | 4.13 | 4.66 | 5.86 | 7.25 | 19.45 | 20.44 | 20.92 | 29.12 | 30.05 | 32.02 |
| ST-GFSL | 2.81 | 3.21 | 4.12 | 4.30 | 5.82 | 7.38 | 19.02 | 20.32 | 20.88 | 28.85 | 29.93 | 31.73 |
| TPB | 2.75 | 3.11 | 3.88 | 4.25 | 5.75 | 6.97 | 19.08 | 20.17 | 20.98 | 28.81 | 29.95 | 31.69 |
| **SSMT (ours)** | **2.66** | **3.01** | **3.80** | **4.14** | **5.60** | **6.82** | **18.58** | **19.45** | **20.34** | **28.31** | **29.35** | **31.02** |

| Methods | Didi-Shenzhen to Didi-Chengdu | | | | | |
|---|---|---|---|---|---|---|
| | MAE | | | RMSE | | |
| | 10 min | 30 min | 60 min | 10 min | 30 min | 60 min |
| DCRNN | 2.68 | 3.19 | 3.41 | 3.55 | 4.05 | 7.62 |
| GWN | 2.79 | 3.05 | 3.49 | 3.41 | 4.12 | 4.77 |
| AdaRNN | 2.64 | 2.91 | 3.35 | 3.33 | 3.97 | 4.56 |
| ST-GFSL | 2.48 | 2.81 | 3.28 | 3.24 | 3.88 | 4.42 |
| TPB | 2.35 | 2.88 | 3.21 | 3.19 | 3.87 | 4.45 |
| **SSMT (ours)** | **2.23** | **2.71** | **3.16** | **3.10** | **3.83** | **4.39** |

### 4.3  Baselines

To demonstrate the superiority and effectiveness of our proposed method in terms of transferability, we use the following baselines: 1) Diffusion Convolutional Recurrent Neural Network (DCRNN) [14] 2) Graph wavenet for deep spatial-temporal graph modeling (GWN) [25] 3) Adaptive learning and forecasting of time series (Ada-RNN) [6], 4) Spatio-Temporal Graph Few-Shot Learning with Cross-City Knowledge Transfer (ST-GFSL) [17], and 5) Cross-city Few-Shot Traffic Forecasting via Traffic Pattern Bank (TPB) [15]. In particular, we compare our model with both transferable and non-transferable methods. The first two models are traditional non-transferable forecasting baselines, whereas the others are transferable baselines including SOTA methods. In this paper, we use the mean absolute error (MAE) and the root mean squared error (RMSE) to evaluate the prediction performance.

### 4.4  Results and Analysis

In Table 2, we present a performance comparison of *SSMT* with other SOTA baselines. Our *SSMT* outperforms all other methods following the same evaluation settings in [17]. We can observe that our proposed *SSMT* outperforms the baselines in both short-term and long-term forecasting. For the METR-LA dataset, our model surpassed the second-best baseline scores, showing enhancements of 2.84% in MAE and 2.44% in RMSE. Similarly, for the PEMSD4 dataset, we observed improvements of 2.90% in MAE and 1.92% in RMSE. Lastly, with the Didi-Chengdu dataset, our gains were 3.14% and 1.50% in MAE and RMSE, respectively. The improved performance demonstrates that the added meta-knowledge from the memory bank and meta-positional encoding indeed enhance the model's forecasting accuracy.

### 4.5   Ablation Study

We delve deeper into analyzing the effectiveness of different modules and the sensitivity of hyperparameters through ablation study. Table 3 demonstrates the effectiveness of the two modules employed in our method. Specifically, we show the experiment on Didi-Chengdu dataset for 1 hour prediction. It clearly illustrates that both strategies are individually effective for our task. Furthermore, combining the results of both methods leads to even greater improvements in performance.

Table 3: Ablation study for the effectiveness of the memory bank and meta-positional encoding (MPE) used in our method.

| Memory | MPE | Error Score (MAE) |
|:---:|:---:|:---:|
| × | × | 4.12% |
| × | ✓ | 3.74% |
| ✓ | × | 3.62% |
| ✓ | ✓ | **3.16%** |

To further investigate the effect of the memory module of *SSMT*, we performed experiments with different memory size settings and presented the corresponding MAE scores for the Didi-Chengdu dataset. The results summarized in Figure 2 demonstrate that when a sufficiently large memory size is used, our *SSMT* consistently generates reliable and plausible outcomes, indicating that the memory module succeeds in transferring informative spatial knowledge from source to the target.

In Figure 3, we examine the impact of the number of tasks on traffic prediction. Our findings indicate that increasing the number of tasks can lead to significant improvements of our model. And, this suggests that diverse temporal tasks have the potential to capture essential temporal patterns specific to the target city.

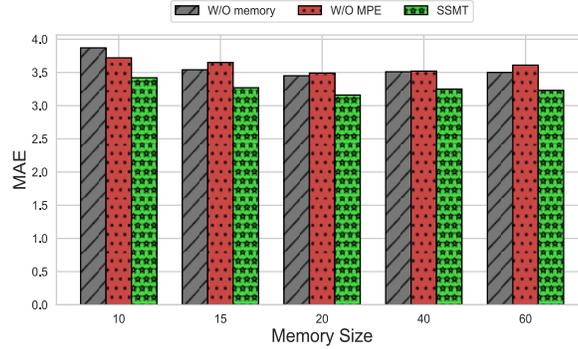

Fig. 2: MAE performance with different memory sizes on PeMSD4 dataset.

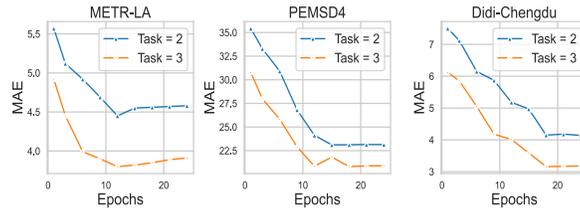

Fig. 3: The performance of *SSMT* vs. the number of tasks.

## 5  Conclusion

In this paper, we present *SSMT*, a transferable model tailored for time series traffic forecasting. During the initial pre-training phase, our proposed model is trained using abundant data from a single source city, while the fine-tuning phase involves refining the pre-trained model using limited data from the target city. We present a carefully designed memory mechanism that stores the diverse patterns from the source city and then retrieve only the target-specific patterns from the memory to predict traffic speed accurately. We further propose a meta-positional encoding that consolidates universal patterns from daily, weekly, and monthly positional encodings. Our experimental results demonstrate the superior transferability, outperforming the SOTA traffic forecasting methods.

**Acknowledgements.** This work was partly supported by Institute for Information & communication Technology Planning & evaluation (IITP) grants funded by the Korean government MSIT: (RS-2022-II221199, RS-2024-00337703, RS-2022-II220688, RS-2019-II190421, RS-2023-00230337, RS-2024-00356293, RS-2022-II221045, and RS-2021-II212068).